\documentclass[sigconf]{acmart}
\setcopyright{none} 
\acmConference[Anonymous Submission to PPML 2021]{ACM ACM Workshop on PRIVACY PRESERVING MACHINE LEARNING}{19 November 2021}{Seoul, South Korea}
\acmYear{2021}
\usepackage{amsmath,amsfonts}
\usepackage{algorithm}
\usepackage{algorithmic}
\usepackage{graphicx}
\usepackage{textcomp}
\usepackage{xcolor}
\usepackage{microtype}
\usepackage{multirow}
\usepackage{subfigure}
\usepackage{booktabs} 
\usepackage{hyperref} 
\usepackage{xspace}
\usepackage{diagbox}          
\usepackage{array}
\usepackage{tabularx}
\usepackage{hyperref} 
\hypersetup{
    colorlinks=true,
    linkcolor=blue,
    citecolor = black,
    urlcolor=teal}

\def\nm{\sigma\xspace}
\def\lr{\eta\xspace}

\newcommand{\latinphrase}[1]{\textit{#1}}

\newcommand{\ie}{\latinphrase{i.e.,}\xspace}
\newcommand{\eg}{\latinphrase{e.g.,}\xspace}

\author{Aman Priyanshu}
\affiliation{%
  \institution{Manipal Institute of Technology}
}
\email{aman.priyanshu@learner.manipal.edu}

\author{Rakshit Naidu}
\affiliation{
    \institution{Carnegie Mellon University}
}
\email{rnemakal@andrew.cmu.edu}

\author{Fatemehsadat Mireshghallah}
\affiliation{
    \institution{University of California, San Diego}
}
\email{fmireshg@eng.ucsd.edu}

\author{Mohammad Malekzadeh}
\affiliation{
    \institution{Imperial College London}
}
\email{m.malekzadeh@imperial.ac.uk}

\begin{document}
\title{Efficient Hyperparameter Optimization for Differentially Private Deep Learning} 

\begin{abstract} 

Tuning the {\em hyperparameters} in the differentially private stochastic gradient descent~(DPSGD) is a fundamental challenge. Unlike the typical SGD, private datasets cannot be used many times for hyperparameter search in DPSGD; \eg via a grid search.  
Therefore, there is an essential need for algorithms that, within a given search space, can  find near-optimal hyperparameters for the best achievable privacy-utility tradeoffs efficiently. We formulate this problem into a general optimization framework for establishing a desirable privacy-utility tradeoff, and systematically study three cost-effective algorithms for being used in the proposed framework: {\em evolutionary}, {\em Bayesian}, and {\em reinforcement learning}. Our experiments, for hyperparameter tuning in DPSGD conducted on MNIST and CIFAR-10 datasets, show that these three algorithms significantly outperform the widely used {\em grid search} baseline. As this paper offers a first-of-a-kind framework for hyperparameter tuning in DPSGD, we discuss existing challenges and open directions for future studies. As we believe our work has implications to be utilized in the pipeline of private deep learning, we open-source our code at~\url{https://github.com/AmanPriyanshu/DP-HyperparamTuning}.
\end{abstract}


\maketitle

\vspace{-1ex}
\section{Introduction}

Deep neural networks~(DNNs)~\cite{goodfellow2016deep} can learn very useful patterns from large multi-dimensional datasets, enabling motivational applications; \eg in health~\cite{ziller2021differentially, dopamine2021}. However, large amounts of training data are required for not only learning the near-optimal DNN parameters for the underlying task, but also for finding the right set of hyperparameters that enable appropriate learning. For a task defined on public datasets, the same data can be reused as many times as we wish. But, as every reuse of the available data comes at a price of some privacy loss, {\em hyperparameter tuning} has been a fundamental challenge for tasks defined on private datasets. 

Differential Privacy (DP)~\cite{dwork2006calibrating} provides strong guarantees for the individuals participating in private datasets. DP restricts the maximum contribution of each sample on the result of a computation on the private dataset. Differentially-private stochastic gradient descent~(DPSGD)~\cite{abadi2016} is a widely accepted algorithm for training DNNs on private datasets, where zero-mean Gaussian noise, with a predefined variance, is added to the clipped gradients computed for each sample in the training dataset at each iteration. Noisy gradients often result in a degraded accuracy for the trained DNN.  


Previous works look at two variants: (1) optimizing privacy parameters of a private model for achieving comparable performance to a non-private model and (2) providing privacy guarantees to reach moderate performance~\cite{van2018three}.
However, in practice, both hyperparameters and privacy parameters need to be optimized within the user-specified privacy budget. Thus, in this paper, we propose a systematic study for learning hyperparameters faster (constrained by a privacy budget) and with less privacy cost through four different optimization algorithms. 

Although there is a wide range of hyperparameters that one can choose from in DPSGD (\eg noise multiplier, clipping factor, batch size, learning rate, etc.), in this paper, we specifically focus on two important hyperparameters: {\em noise multiplier $\nm$} (\ie the standard deviation of the Gaussian noise) and {\em learning rate $\lr$}. We optimize for these two parameters specifically as the epsilon ($\epsilon$ i.e. privacy leakage) and validation loss (minimizing the validation loss) are highly dependent on the chosen values for $\nm$ and $\lr$, respectively. To this end, we study three cost-effective algorithms: {\em evolutionary}, {\em Bayesian}, and {\em reinforcement learning} and compare the results with the {\em grid search} base-line. We also display consistent results across these techniques and provide insights on which algorithm could pave the path towards better hyperparameter tuning in DPSGD. We also open-source our code to enable future practitioners' optimize for specific hyperparameters, according to their requirements.

\section{Related Work}
The most widely used methods for hyperparameter tuning in deep learning are manual search, random search, and grid search~\cite{tramer2021differentially}. Manual search refers to the manual tuning of hyperparameters by individuals experimenting on a deep neural network. This method is frequently used due to prior experiences and intuition. On the other hand, random search provides a path towards hyperparameter space exploration. However, it is non-exhaustive and may not be able to discover high-performing hyperparameters. Thus, grid search is utilized to provide a sufficient exploration within a restricted search space.  Although, due to its non-adaptive nature (i.e., the hyper-parameter sets selected to be evaluated are not selected using already available results), it utilizes abundant resources and requires significant computational time.

Not all common practices in training deep models are always directly applicable when we apply DPSGD. For instance, \cite{papernot2020tempered} shows that while  {\em ReLU} is the most common activation function in conventional deep learning, for DPSGD a bounded activation function, such as {\em tanh} or tempered {\em sigmoid}, is more efficient. Also, \cite{van2018three} argues that while SGD hyperparameters and DP parameters are considered independent in the original DPSGD paper~\cite{abadi2016}, this is not a reasonable assumption. For example, by choosing a smaller  batch size, we can achieve better privacy, but to keep accuracy well, we thus need to also reduce the learning rate accordingly. However, smaller learning rates usually slow down the convergence~\cite{bengio2012practical}, thus we need more epochs and hence more privacy loss in DPSGD. Therefore, \cite{van2018three} suggests using a public dataset first, to find an appropriate DNN architecture with an optimized set of hyperparameters, and then train the model on the private dataset. However, \cite{van2018three} did not propose any method for searching over DPSGD hyperparameters.

Model selection in multivariate linear regression under the constraint of differential privacy is studied in~\cite{lei2016differentially}; based on penalized least squares and likelihood. Especially, \cite{lei2016differentially} reports that under differential privacy, the procedure of model selection becomes more sensitive to the tuning parameters. Moreover, the appropriate choice of tuning parameters requires some additional information in the data, and it is mentioned as a future topic in~\cite{lei2016differentially} to develop differentially private methods to estimate these hyperparameters. 
\cite{pichapati2019adaclip} introduces AdaCliP, which achieves the same privacy guarantee with much less added noise by using coordinate-wise adaptive clipping of the gradient. As the convergence of DPSGD depends on the variance of the gradient, AdaCliP also improves the accuracy of the trained model. While such an adaptive clipping provides better tradeoffs, it needs to estimate the variance and thus introduces four new hyperparameters by itself, making our problem more complicated. Similarly, \cite{andrew2019differentially} introduces a method for adaptively tuning the clipping threshold to track a given ``quantile'' of the update norm distribution during training. Again, this method also needs to tune a new hyperparameter in the range of $[0,1]$. 

The DPareto algorithm is proposed in~\cite{avent2019automatic}, where Bayesian optimization is used for hyperparameter tuning. The paper describes the Pareto Front and empirically validates its application using different neural network architectures across two datasets. Their study uses the multi-objective Bayesian optimizer to find the best hyperparameters, utilizing hypervolume to find the relative merit of different objectives. On the other hand, we use a single-objective Bayesian optimizer for our study, which reduces computational costs and aims to optimize the reward function defined by us. 

\cite{yazdanbakhsh2018releq} uses reinforcement learning to efficiently tune hyperparameters needed for quantization of deep neural networks and find the bit-widths for weights of each layer that would provide optimal computation-accuracy trade-off. 

\begin{figure}[t]
    \centering
     \includegraphics[width=.65\linewidth]{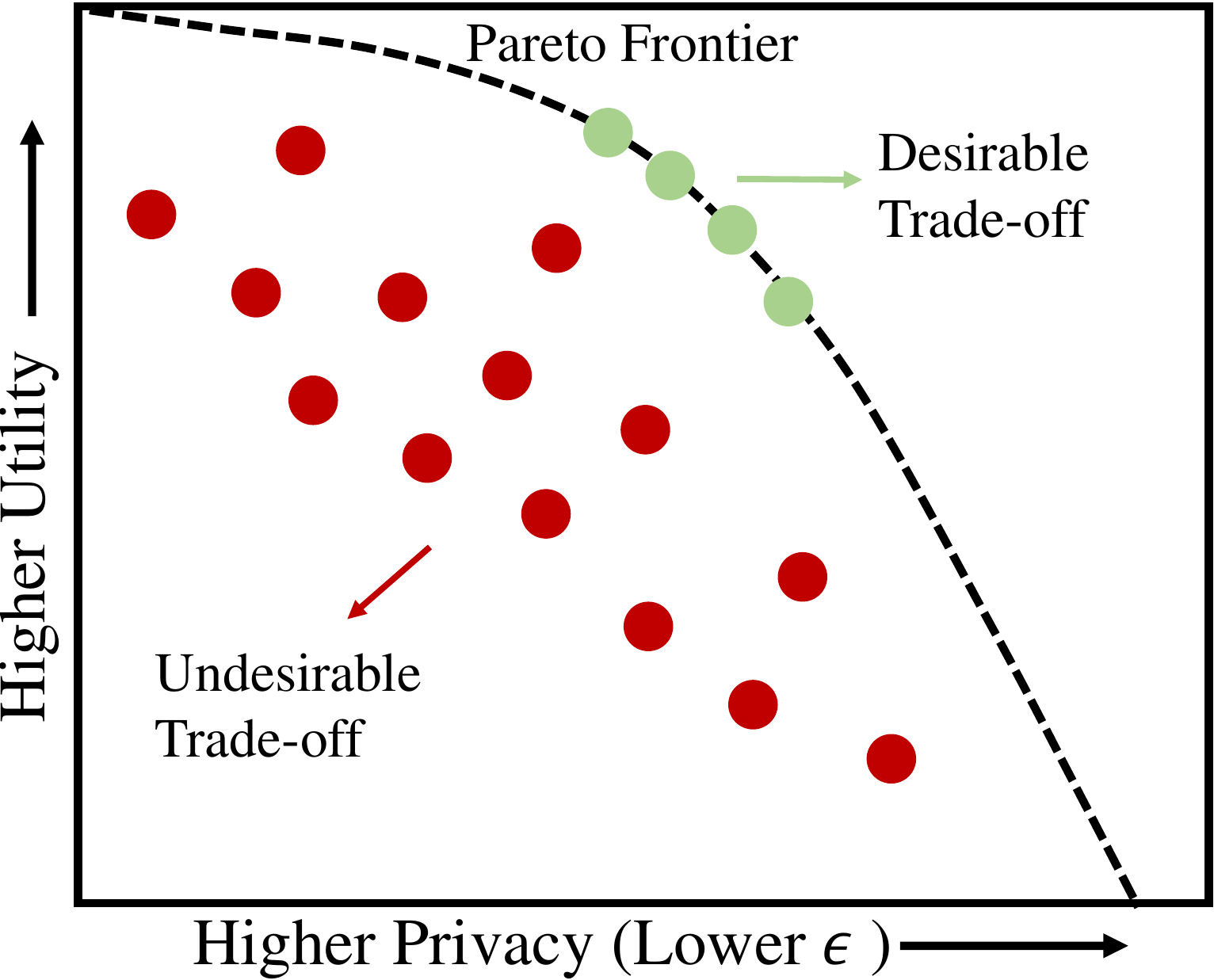}
     \caption{The Pareto frontier of potential privacy-utility trade-offs. Each point~(green/red) shows the  trade-off by choosing a specific set of values for the hyperparameters. }
     
     \label{fig:pareto}
    \vspace{-2ex}
\end{figure}

\section{Methodologies}

\subsection{Problem Formulation}

We consider the problem of training a DNN with a fixed architecture (i.e., the number, type, and size of each layer) using DPSGD. Let $D_{train}$ denotes the training set and $D_{valid}$ denote the validation set. Let $\mathcal{H} = \{h_1, \dots, h_N\}$ denotes the set of $N$ {\em hyperparameters} that are used during training on $D_{train}$ and have impact on both {\em validation loss} ($val\_loss$) and {\em privacy loss} in DP ($\epsilon$), on $D_{valid}$. To provide a general but customizable framework, we define {\em reward} as a weighted linear combination of $val\_loss$ and $\epsilon$: 
\begin{equation}
    reward = \alpha^{U}.(e^{-val\_loss} )+\alpha^{P}.(e^{-\epsilon}).
    \label{equ1}
\end{equation}
We use regularizers $\alpha^{U}$ and $\alpha^{P} \in [0, 1]$ to control the importance of utility and privacy, respectively (to control the privacy-utility trade-off). In our proposed framework, we first set these $\alpha$ regularizers, and then start searching for the optimal hyperparameters in $\mathcal{H}$ using the algorithms explained in the following section. In this paper, we consider $\mathcal{H}$ = \{$\nm$, $\lr$\},  where $\nm$ denotes the noise multiplier and $\lr$ denotes the learning rate; in DPSGD. Our aim for the following experiments remains to optimize the reward given by Equation~\eqref{equ1}. Notice that, in practice, the value of $\alpha^{U}$ and $\alpha^{P}$ depends on the requirements of the underlying task.


\subsection{Evolutionary optimization}
Evolutionary optimization algorithms provide an opportunity to explore as well as exploit the hyperparameter search space \cite{8297018, 10.1145/2834892.2834896}. Its random initialization and mutational attributes allow it to take advantage of the random search optimization algorithm. In contrast, its adaptive nature enables it to exploit critical values, which give better results. In our implementation, we encoded each hyperparameter as a gene, a set of hyperparameters made up the genome of an individual, i.e., the experiment. The range and precision for each hyperparameter are predetermined, allowing the optimization algorithm to search within a limited search space. The initial population is determined by random sampling of hyperparameters from this search space. Once aggregated, the population is trained, and a fitness score or reward is measured using \ref{equ1}. Subsequently, each generation is formed using selection, cross-over, and mutation based on the individuals with the highest fitness from the previous generation. The methodology can be optimized for high exploitation, thereby reducing resource wastage \cite{10.1145/2834892.2834896}.

\begin{table}[t]
\small
\caption{Comparison of different methods based on the best-achieved reward and the average time required to attain this reward for the search space on (A) CIFAR-10 and (B) MNIST.}
\label{tab:result_1}
\begin{tabular}{@{}lcccc@{}}
\toprule
Method        & Time & Best Reward & Accuracy & Epsilon ($\epsilon$) \\ 
& (in hours) & (in \%) & (in \%) & \\\toprule
\multicolumn{5}{c}{(A) CIFAR-10} \\\toprule
Grid Search   & 150.020 & 51.406 & \textbf{44.936} & 0.600 \\
Evolutionary  & \textbf{11.064} & 52.044 & 37.999 & 0.599   \\
Bayesian      & 49.636 & 51.846 & 43.864 & \textbf{0.581} \\
Reinforcement & 52.971 & \textbf{52.398} & 44.884 & 0.590 \\ \bottomrule
\multicolumn{5}{c}{(B) MNIST} \\\toprule

%
Grid Search   & 43.712 & 72.260 & \textbf{89.133} & 0.683  \\
Evolutionary  & 5.250 & 72.615 & 73.745 & \textbf{0.175}  \\
Bayesian      & \textbf{2.853} & 73.385 & 81.562 & 0.349  \\
Reinforcement & 31.165 & \textbf{74.906} & 75.022 & 0.240 \\ \bottomrule
\end{tabular}
\end{table}

\subsection{Bayesian Optimization}
Bayesian optimization treats neural network training and performance as a black-box function. It combines prior experience with the black-box neural network with sample information to approximate the function distribution using the Bayesian formula \cite{WU201926}. Based on this estimation of the function distribution, optimal values can be extrapolated. The estimate distribution is effectively a probabilistic model for the function, which exploits this model to decide where to explore the function next while integrating out uncertainty \cite{NIPS2012_05311655}. This methodology allows one to find the minima of complex non-convex functions with relatively few evaluations. However, this is only due to our assumption that the function is drawn from a Gaussian process prior. Our experiments utilized Hyperopt, a Sequential Model-Based Optimization (SMBO) that provides high performance at a low computational budget \cite{Bergstra_hyperopt}. 

\subsection{Reinforcement Learning}
Evolutionary optimization and Bayesian optimization provide both adequate methodologies for search space exploration and exploitation. However, this classical problem can also be dealt with by reinforcement learning. In our application of this method, we begin by initializing a regression network capable of estimating the reward output of training on a particular set of hyperparameters. We start by sampling a random collection of hyperparameters used to train the DPSGD model and obtain the reward to fit the regression network. We then proceed to extract the estimated reward of the entire search space for our hyperparameter tuning. The best-performing hyperparameters are obtained from this estimation. These hyperparameters are mutated to hyperparameters in near proximity to them for the next episode, thereby allowing the model to exploit values that may give high performance. Subsequently, in the following episodes, we select a certain percent of experiments based on the reward estimate of the regression network; in contrast, the others continue to be randomly sampled. This is determined based on the epsilon-decreasing strategy, where the value of exploration-exploitation-epsilon decreases as the experiment progresses. This methodology allows us to estimate the hyperparameter-reward function and verify the proximal search space of high-performing hyperparameters, giving us generalized results.

\section{Evaluation}
For the experiments in this section, we used a Tesla P100 16GB as GPU, with 13GB RAM Intel Xeon as CPU for our experiments. Note that the random seed is fixed across all experiments for uniformity and reproducibility purposes. In the rest of this section, we will discuss the benchmarks used and the results of each experiment.  
\subsection{Benchmarks}

To assess and analyze the effectiveness of optimization algorithms across both CIFAR-10 and MNIST datasets, we use Grid-Search on a similar search complexity as the other methods. We display the computational time taken, best reward achieved, and its respective accuracy and epsilon value in Table \ref{tab:result_1}. Grid search displays a poor understanding of the epsilon-accuracy as it is not adaptive in nature. It achieves a reward of 72.2\% and 51.4\% on the MNIST and CIFAR-10 datasets, respectively.

\subsection{Optimization Algorithms}

As described in earlier sections, we ran our experiments over three distinct optimization algorithms. Here, we observe that although Reinforcement Learning provides the highest performance, it comes at the expense of computational time. On the other hand, Evolutionary Algorithms and Bayesian Optimization provide consistent results with respect to computational time and performance.

Additionally, we can see from Table \ref{tab:result_1} that although Grid Search returns a highly accurate model, it is compensated by the high privacy leakage that occurs due to it. Contrary, adaptive optimization algorithms can leverage previous samples to search for a better privacy-utility tradeoff, allowing them to achieve high rewards. We see that Evolutionary Optimization achieves the lowest epsilon value for the MNIST dataset. In contrast, Bayesian Optimization achieves the same for the CIFAR-10 dataset.

\begin{figure}[t]
    \centering
     \includegraphics[width=0.92\linewidth]{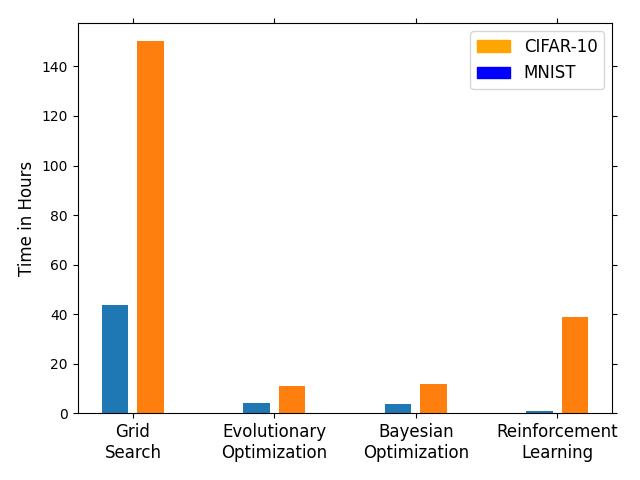}
     \caption{The time taken by different methodologies to achieve reward greater than or equal to baseline rewards. 
     \label{fig:time_histo}}

    \centering
     \includegraphics[width=0.92\linewidth]{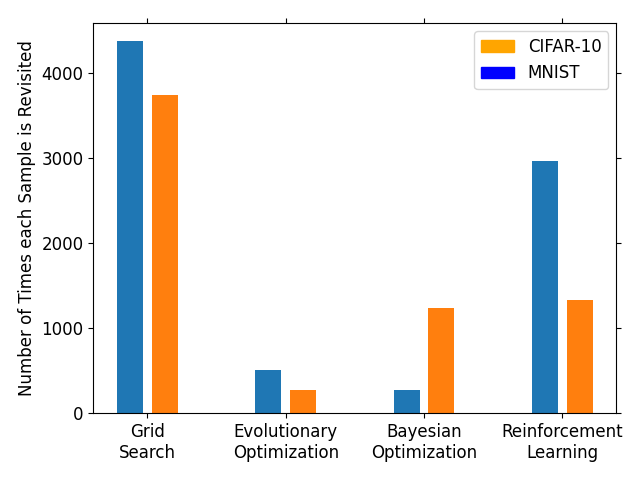}
     \caption{The number of times each sample is re-visited before the best reward is achieved. 
     \label{fig:num_histo}}
    \vspace{-2ex}
\end{figure}

\begin{figure*}[t]
    \centering
     \includegraphics[width=\linewidth]{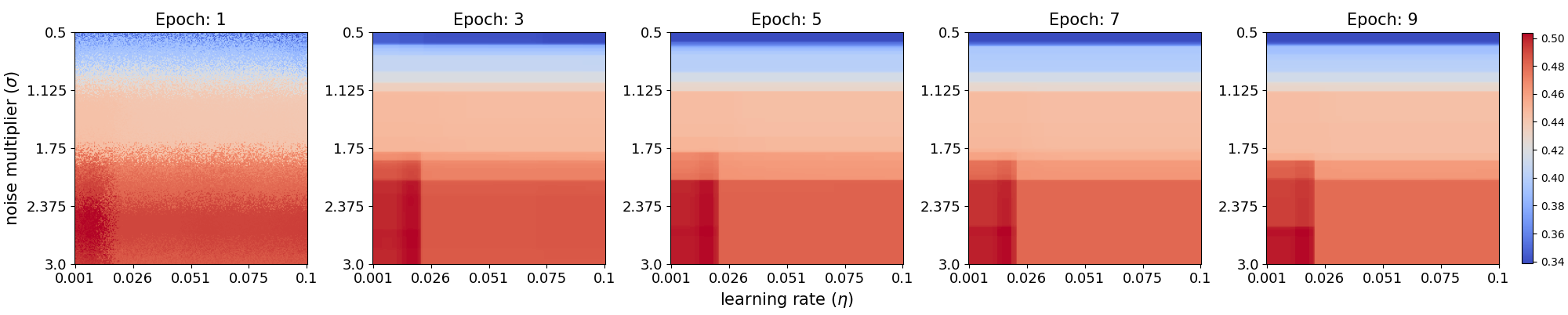}
     \includegraphics[width=\linewidth]{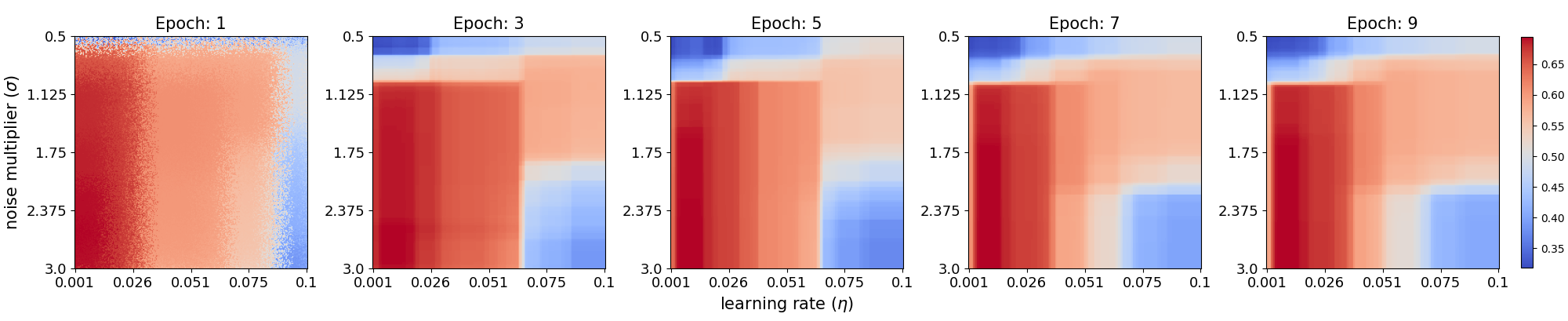}
     \caption{Convergence of best reward estimate of the ANN model, based on input $noise\_multiplier$ ($\nm$, Y-Axis) and $learning\_rate$ ($\lr$, X-Axis). We display model convergence for the CIFAR-10 dataset on top and the MNIST dataset at the bottom.
    }
     \label{fig:cifar_RL}
    \vspace{-2ex}
\end{figure*}
\subsection{Evaluating Computation Time for Satisfactory Reward}

Although finding the best reward is our goal, we also evaluate the computational time required by each algorithm to achieve the maximum reward attained by Grid Search. The time consumed is calculated based on the time taken for an optimization algorithm to achieve a reward equal to or greater than the baseline reward. Here, baseline reward refers to the highest reward achieved by the Grid Search algorithm. We display these results in Figure \ref{fig:time_histo}. It can clearly be seen that grid search is much more time-consuming in nature, compared to other optimization algorithms.

From Figure \ref{fig:time_histo}, we can observe that Bayesian and evolutionary optimization algorithms display consistency and uniformity across datasets. Whereas, reinforcement-learning-based optimization fails to generalize the time taken to achieve a given reward. This is due to the exploration-exploitation trade-off mechanism applied by the aforementioned method. An appropriate exploration-exploitation-epsilon must be selected if the user intends to have a highly exploitative model for optimization.


\subsection{Evaluation Sample-Specific Privacy}

In this subsection, we look at sample-specific privacy. We count the total number of times that each sample has revisited for each algorithm during optimization. As training on any sample at a given time can lead to sensitive information leakage, we consider this an essential feature for privacy evaluation. In Figure \ref{fig:num_histo}, we display a bar graph representing the number of times any sample in the training set is visited before every model achieves its respective highest reward. These evaluations continue to display the impractical nature of Grid Search and validate its high privacy leakage. On the other hand, Bayesian and Evolutionary Optimization continues to give expected results compared to the prior. The Reinforcement Learning approach again displays high variance across datasets, giving it a weaker generalization capacity.

\subsection{Learning and Convergence Analysis of Reinforcement Learning Approach}

We further study the behavior of the reinforcement learning approach in our analysis through Figure~\ref{fig:cifar_RL}. As the model learns over newer data every epoch, we can observe that we continue to adapt the expected reward accordingly. The model makes more resolute changes to the reward estimate nearest to the previous global maxima. This allows the model to exploit better performing hyperparameter values, allowing it to restrict search within a high-performing area. However, this also supplements that as the number of epochs increases, estimates which remain unexplored do not change in value. Therefore, an experimentally robust exploration-exploitation-epsilon must be selected for generalized results.

On comparing the two datasets, we can see the more complex nature of the search space for the MNIST dataset. This can be attributed to the simple complexity of the dataset, which allows learning over different hyperparameters. However, the CIFAR-10 dataset is much more complex, leading to only the most optimal hyperparameters being highlighted.


\section{Conclusion and Future Work}

In this paper, we discussed different methodologies for hyperparameter tuning for the private training of deep neural networks using DPSGD algorithm. We proposed a novel, customizable reward function that allows users to define a single objective function for establishing their desired privacy-utility tradeoff.  We quantified, compared, and analyzed the methods of grid search (as the baseline), Bayesian optimization, evolutionary optimization, and reinforcement learning, across two datasets,  CIFAR-10, and  MNIST. We observed that Bayesian and evolutionary optimization behave similarly in terms of the privacy-utility trade-off point they provide, and how efficiently they find it. Reinforcement learning, however, provides a more desirable trade-off but with varying efficiencies across datasets. All three methods perform much better than the baseline grid search algorithm. We believe that our work serves as a valuable resource for privacy-preserving ML practitioners, developers, and researchers for hyperparameter tuning. 

For future work, one can use our proposed method alongside that of~\cite{lee2018concentrated, chen2020stochastic}, where a portion of the privacy budget is allocated to finding the appropriate learning rate on the private dataset. Another direction is to extend our proposed method to tune other hyperparameters in DPSGD, and even the network architecture and non-linear activation functions that are used.  


\subsection*{Acknowledgement} 
Mohammad Malekzadeh was partially supported by the UK EPSRC (grant no. EP/T023600/1) within the CHIST-ERA program.
\bibliographystyle{ACM-Reference-Format}
\bibliography{refs} 


\begin{thebibliography}{00}


\ifx \showCODEN    \undefined \def \showCODEN     #1{\unskip}     \fi
\ifx \showDOI      \undefined \def \showDOI       #1{#1}\fi
\ifx \showISBNx    \undefined \def \showISBNx     #1{\unskip}     \fi
\ifx \showISBNxiii \undefined \def \showISBNxiii  #1{\unskip}     \fi
\ifx \showISSN     \undefined \def \showISSN      #1{\unskip}     \fi
\ifx \showLCCN     \undefined \def \showLCCN      #1{\unskip}     \fi
\ifx \shownote     \undefined \def \shownote      #1{#1}          \fi
\ifx \showarticletitle \undefined \def \showarticletitle #1{#1}   \fi
\ifx \showURL      \undefined \def \showURL       {\relax}        \fi
\providecommand\bibfield[2]{#2}
\providecommand\bibinfo[2]{#2}
\providecommand\natexlab[1]{#1}
\providecommand\showeprint[2][]{arXiv:#2}

\bibitem[\protect\citeauthoryear{Abadi, Chu, Goodfellow, McMahan, Mironov,
  Talwar, and Zhang}{Abadi et~al\mbox{.}}{2016}]%
        {abadi2016}
\bibfield{author}{\bibinfo{person}{Martin Abadi}, \bibinfo{person}{Andy Chu},
  \bibinfo{person}{Ian Goodfellow}, \bibinfo{person}{H~Brendan McMahan},
  \bibinfo{person}{Ilya Mironov}, \bibinfo{person}{Kunal Talwar}, {and}
  \bibinfo{person}{Li Zhang}.} \bibinfo{year}{2016}\natexlab{}.
\newblock \showarticletitle{Deep learning with differential privacy}. In
  \bibinfo{booktitle}{{\em Proceedings of the 2016 ACM SIGSAC conference on
  computer and communications security}}. \bibinfo{pages}{308--318}.
\newblock


\bibitem[\protect\citeauthoryear{Andrew, Thakkar, McMahan, and
  Ramaswamy}{Andrew et~al\mbox{.}}{2019}]%
        {andrew2019differentially}
\bibfield{author}{\bibinfo{person}{Galen Andrew}, \bibinfo{person}{Om Thakkar},
  \bibinfo{person}{H~Brendan McMahan}, {and} \bibinfo{person}{Swaroop
  Ramaswamy}.} \bibinfo{year}{2019}\natexlab{}.
\newblock \showarticletitle{Differentially private learning with adaptive
  clipping}.
\newblock \bibinfo{journal}{{\em arXiv preprint arXiv:1905.03871\/}}
  (\bibinfo{year}{2019}).
\newblock


\bibitem[\protect\citeauthoryear{Avent, González, Diethe, Paleyes, and
  Balle}{Avent et~al\mbox{.}}{2020}]%
        {avent2019automatic}
\bibfield{author}{\bibinfo{person}{Brendan Avent}, \bibinfo{person}{Javier
  González}, \bibinfo{person}{Tom Diethe}, \bibinfo{person}{Andrei Paleyes},
  {and} \bibinfo{person}{Borja Balle}.} \bibinfo{year}{2020}\natexlab{}.
\newblock \showarticletitle{Automatic Discovery of Privacy–Utility Pareto
  Fronts}.
\newblock \bibinfo{journal}{{\em Proceedings on Privacy Enhancing
  Technologies\/}} \bibinfo{volume}{2020}, \bibinfo{number}{4}
  (\bibinfo{year}{2020}), \bibinfo{pages}{5--23}.
\newblock


\bibitem[\protect\citeauthoryear{Bengio}{Bengio}{2012}]%
        {bengio2012practical}
\bibfield{author}{\bibinfo{person}{Yoshua Bengio}.}
  \bibinfo{year}{2012}\natexlab{}.
\newblock \showarticletitle{Practical recommendations for gradient-based
  training of deep architectures}.
\newblock In \bibinfo{booktitle}{{\em Neural networks: Tricks of the trade}}.
  \bibinfo{publisher}{Springer}, \bibinfo{pages}{437--478}.
\newblock


\bibitem[\protect\citeauthoryear{Bergstra, Yamins, and Cox}{Bergstra
  et~al\mbox{.}}{}]%
        {Bergstra_hyperopt}
\bibfield{author}{\bibinfo{person}{James Bergstra}, \bibinfo{person}{Dan
  Yamins}, {and} \bibinfo{person}{David~D. Cox}.}
\newblock \bibinfo{title}{Hyperopt: A Python Library for Optimizing the
  Hyperparameters of Machine Learning Algorithms}.
\newblock   (\bibinfo{year}{????}).
\newblock


\bibitem[\protect\citeauthoryear{Bochinski, Senst, and Sikora}{Bochinski
  et~al\mbox{.}}{2017}]%
        {8297018}
\bibfield{author}{\bibinfo{person}{Erik Bochinski}, \bibinfo{person}{Tobias
  Senst}, {and} \bibinfo{person}{Thomas Sikora}.}
  \bibinfo{year}{2017}\natexlab{}.
\newblock \showarticletitle{Hyper-parameter optimization for convolutional
  neural network committees based on evolutionary algorithms}. In
  \bibinfo{booktitle}{{\em 2017 IEEE International Conference on Image
  Processing (ICIP)}}. \bibinfo{pages}{3924--3928}.
\newblock
\showDOI{%
\url{https://doi.org/10.1109/ICIP.2017.8297018}}


\bibitem[\protect\citeauthoryear{Chen and Lee}{Chen and Lee}{2020}]%
        {chen2020stochastic}
\bibfield{author}{\bibinfo{person}{Chen Chen} {and} \bibinfo{person}{Jaewoo
  Lee}.} \bibinfo{year}{2020}\natexlab{}.
\newblock \showarticletitle{Stochastic adaptive line search for differentially
  private optimization}. In \bibinfo{booktitle}{{\em 2020 IEEE International
  Conference on Big Data (Big Data)}}. IEEE, \bibinfo{pages}{1011--1020}.
\newblock


\bibitem[\protect\citeauthoryear{Dwork, McSherry, Nissim, and Smith}{Dwork
  et~al\mbox{.}}{2006}]%
        {dwork2006calibrating}
\bibfield{author}{\bibinfo{person}{Cynthia Dwork}, \bibinfo{person}{Frank
  McSherry}, \bibinfo{person}{Kobbi Nissim}, {and} \bibinfo{person}{Adam
  Smith}.} \bibinfo{year}{2006}\natexlab{}.
\newblock \showarticletitle{Calibrating noise to sensitivity in private data
  analysis}. In \bibinfo{booktitle}{{\em Theory of cryptography conference}}.
  Springer, \bibinfo{pages}{265--284}.
\newblock


\bibitem[\protect\citeauthoryear{Goodfellow, Bengio, and Courville}{Goodfellow
  et~al\mbox{.}}{2016}]%
        {goodfellow2016deep}
\bibfield{author}{\bibinfo{person}{Ian Goodfellow}, \bibinfo{person}{Yoshua
  Bengio}, {and} \bibinfo{person}{Aaron Courville}.}
  \bibinfo{year}{2016}\natexlab{}.
\newblock \bibinfo{booktitle}{{\em Deep learning}}.
\newblock \bibinfo{publisher}{MIT press}.
\newblock


\bibitem[\protect\citeauthoryear{Lee and Kifer}{Lee and Kifer}{2018}]%
        {lee2018concentrated}
\bibfield{author}{\bibinfo{person}{Jaewoo Lee} {and} \bibinfo{person}{Daniel
  Kifer}.} \bibinfo{year}{2018}\natexlab{}.
\newblock \showarticletitle{Concentrated differentially private gradient
  descent with adaptive per-iteration privacy budget}. In
  \bibinfo{booktitle}{{\em Proceedings of the 24th ACM SIGKDD International
  Conference on Knowledge Discovery \& Data Mining}}.
  \bibinfo{pages}{1656--1665}.
\newblock


\bibitem[\protect\citeauthoryear{Lei, Charest, Slavkovic, Smith, and
  Fienberg}{Lei et~al\mbox{.}}{2018}]%
        {lei2016differentially}
\bibfield{author}{\bibinfo{person}{Jing Lei}, \bibinfo{person}{Anne-Sophie
  Charest}, \bibinfo{person}{Aleksandra Slavkovic}, \bibinfo{person}{Adam
  Smith}, {and} \bibinfo{person}{Stephen Fienberg}.}
  \bibinfo{year}{2018}\natexlab{}.
\newblock \showarticletitle{Differentially private model selection with
  penalized and constrained likelihood}.
\newblock \bibinfo{journal}{{\em Journal of the Royal Statistical Society:
  Series A (Statistics in Society)\/}} \bibinfo{volume}{181},
  \bibinfo{number}{3} (\bibinfo{year}{2018}), \bibinfo{pages}{609--633}.
\newblock


\bibitem[\protect\citeauthoryear{Malekzadeh, Hasircioglu, Mital, Katarya,
  Ozfatura, and Gündüz}{Malekzadeh et~al\mbox{.}}{2021}]%
        {dopamine2021}
\bibfield{author}{\bibinfo{person}{Mohammad Malekzadeh}, \bibinfo{person}{Burak
  Hasircioglu}, \bibinfo{person}{Nitish Mital}, \bibinfo{person}{Kunal
  Katarya}, \bibinfo{person}{Mehmet~Emre Ozfatura}, {and}
  \bibinfo{person}{Deniz Gündüz}.} \bibinfo{year}{2021}\natexlab{}.
\newblock \showarticletitle{Dopamine: Differentially Private Federated Learning
  on Medical Data}.
\newblock \bibinfo{journal}{{\em The Second AAAI Workshop on Privacy-Preserving
  Artificial Intelligence (PPAI-21)\/}} (\bibinfo{year}{2021}).
\newblock


\bibitem[\protect\citeauthoryear{Papernot, Thakurta, Song, Chien, and
  Erlingsson}{Papernot et~al\mbox{.}}{2021}]%
        {papernot2020tempered}
\bibfield{author}{\bibinfo{person}{Nicolas Papernot},
  \bibinfo{person}{Abhradeep Thakurta}, \bibinfo{person}{Shuang Song},
  \bibinfo{person}{Steve Chien}, {and} \bibinfo{person}{Ulfar Erlingsson}.}
  \bibinfo{year}{2021}\natexlab{}.
\newblock \showarticletitle{Tempered Sigmoid Activations for Deep Learning with
  Differential Privacy}. In \bibinfo{booktitle}{{\em Proceedings of the AAAI
  Conference on Artificial Intelligence}}.
\newblock


\bibitem[\protect\citeauthoryear{Pichapati, Suresh, Yu, Reddi, and
  Kumar}{Pichapati et~al\mbox{.}}{2019}]%
        {pichapati2019adaclip}
\bibfield{author}{\bibinfo{person}{Venkatadheeraj Pichapati},
  \bibinfo{person}{Ananda~Theertha Suresh}, \bibinfo{person}{Felix~X Yu},
  \bibinfo{person}{Sashank~J Reddi}, {and} \bibinfo{person}{Sanjiv Kumar}.}
  \bibinfo{year}{2019}\natexlab{}.
\newblock \showarticletitle{AdaCliP: Adaptive clipping for private SGD}.
\newblock \bibinfo{journal}{{\em arXiv preprint arXiv:1908.07643\/}}
  (\bibinfo{year}{2019}).
\newblock


\bibitem[\protect\citeauthoryear{Snoek, Larochelle, and Adams}{Snoek
  et~al\mbox{.}}{2012}]%
        {NIPS2012_05311655}
\bibfield{author}{\bibinfo{person}{Jasper Snoek}, \bibinfo{person}{Hugo
  Larochelle}, {and} \bibinfo{person}{Ryan~P Adams}.}
  \bibinfo{year}{2012}\natexlab{}.
\newblock \showarticletitle{Practical Bayesian Optimization of Machine Learning
  Algorithms}. In \bibinfo{booktitle}{{\em Advances in Neural Information
  Processing Systems}}, \bibfield{editor}{\bibinfo{person}{F.~Pereira},
  \bibinfo{person}{C.~J.~C. Burges}, \bibinfo{person}{L.~Bottou}, {and}
  \bibinfo{person}{K.~Q. Weinberger}} (Eds.), Vol.~\bibinfo{volume}{25}.
  \bibinfo{publisher}{Curran Associates, Inc.}
\newblock
\showURL{%
\url{https://proceedings.neurips.cc/paper/2012/file/05311655a15b75fab86956663e1819cd-Paper.pdf}}


\bibitem[\protect\citeauthoryear{Tramer and Boneh}{Tramer and Boneh}{2021}]%
        {tramer2021differentially}
\bibfield{author}{\bibinfo{person}{Florian Tramer} {and} \bibinfo{person}{Dan
  Boneh}.} \bibinfo{year}{2021}\natexlab{}.
\newblock \showarticletitle{Differentially Private Learning Needs Better
  Features (or Much More Data)}. In \bibinfo{booktitle}{{\em International
  Conference on Learning Representations (ICLR)}}.
\newblock


\bibitem[\protect\citeauthoryear{van~der Veen, Seggers, Bloem, and
  Patrini}{van~der Veen et~al\mbox{.}}{2018}]%
        {van2018three}
\bibfield{author}{\bibinfo{person}{Koen~Lennart van~der Veen},
  \bibinfo{person}{Ruben Seggers}, \bibinfo{person}{Peter Bloem}, {and}
  \bibinfo{person}{Giorgio Patrini}.} \bibinfo{year}{2018}\natexlab{}.
\newblock \showarticletitle{Three tools for practical differential privacy}.
\newblock \bibinfo{journal}{{\em arXiv preprint arXiv:1812.02890\/}}
  (\bibinfo{year}{2018}).
\newblock


\bibitem[\protect\citeauthoryear{Wu, Chen, Zhang, Xiong, Lei, and Deng}{Wu
  et~al\mbox{.}}{2019}]%
        {WU201926}
\bibfield{author}{\bibinfo{person}{Jia Wu}, \bibinfo{person}{Xiu-Yun Chen},
  \bibinfo{person}{Hao Zhang}, \bibinfo{person}{Li-Dong Xiong},
  \bibinfo{person}{Hang Lei}, {and} \bibinfo{person}{Si-Hao Deng}.}
  \bibinfo{year}{2019}\natexlab{}.
\newblock \showarticletitle{Hyperparameter Optimization for Machine Learning
  Models Based on Bayesian Optimizationb}.
\newblock \bibinfo{journal}{{\em Journal of Electronic Science and
  Technology\/}} \bibinfo{volume}{17}, \bibinfo{number}{1}
  (\bibinfo{year}{2019}), \bibinfo{pages}{26--40}.
\newblock
\showISSN{1674-862X}
\showDOI{%
\url{https://doi.org/10.11989/JEST.1674-862X.80904120}}


\bibitem[\protect\citeauthoryear{Yazdanbakhsh, Elthakeb, Pilligundla,
  Mireshghallah, and Esmaeilzadeh}{Yazdanbakhsh et~al\mbox{.}}{2018}]%
        {yazdanbakhsh2018releq}
\bibfield{author}{\bibinfo{person}{Amir Yazdanbakhsh}, \bibinfo{person}{Ahmed~T
  Elthakeb}, \bibinfo{person}{Prannoy Pilligundla}, \bibinfo{person}{F
  Mireshghallah}, {and} \bibinfo{person}{Hadi Esmaeilzadeh}.}
  \bibinfo{year}{2018}\natexlab{}.
\newblock \showarticletitle{Releq: An automatic reinforcement learning approach
  for deep quantization of neural networks}.
\newblock \bibinfo{journal}{{\em arXiv preprint arXiv:1811.01704\/}}
  \bibinfo{volume}{1}, \bibinfo{number}{2} (\bibinfo{year}{2018}).
\newblock


\bibitem[\protect\citeauthoryear{Young, Rose, Karnowski, Lim, and Patton}{Young
  et~al\mbox{.}}{2015}]%
        {10.1145/2834892.2834896}
\bibfield{author}{\bibinfo{person}{Steven~R. Young}, \bibinfo{person}{Derek~C.
  Rose}, \bibinfo{person}{Thomas~P. Karnowski}, \bibinfo{person}{Seung-Hwan
  Lim}, {and} \bibinfo{person}{Robert~M. Patton}.}
  \bibinfo{year}{2015}\natexlab{}.
\newblock \showarticletitle{Optimizing Deep Learning Hyper-Parameters through
  an Evolutionary Algorithm}. In \bibinfo{booktitle}{{\em Proceedings of the
  Workshop on Machine Learning in High-Performance Computing Environments}}
  {\em (\bibinfo{series}{MLHPC '15})}. \bibinfo{publisher}{Association for
  Computing Machinery}, \bibinfo{address}{New York, NY, USA}, Article
  \bibinfo{articleno}{4}, \bibinfo{numpages}{5}~pages.
\newblock
\showISBNx{9781450340069}
\showDOI{%
\url{https://doi.org/10.1145/2834892.2834896}}


\bibitem[\protect\citeauthoryear{Ziller, Usynin, Remerscheid, Knolle, Makowski,
  Braren, Rueckert, and Kaissis}{Ziller et~al\mbox{.}}{2021}]%
        {ziller2021differentially}
\bibfield{author}{\bibinfo{person}{Alexander Ziller}, \bibinfo{person}{Dmitrii
  Usynin}, \bibinfo{person}{Nicolas Remerscheid}, \bibinfo{person}{Moritz
  Knolle}, \bibinfo{person}{Marcus Makowski}, \bibinfo{person}{Rickmer Braren},
  \bibinfo{person}{Daniel Rueckert}, {and} \bibinfo{person}{Georgios Kaissis}.}
  \bibinfo{year}{2021}\natexlab{}.
\newblock \showarticletitle{Differentially private federated deep learning for
  multi-site medical image segmentation}.
\newblock \bibinfo{journal}{{\em arXiv preprint arXiv:2107.02586\/}}
  (\bibinfo{year}{2021}).
\newblock


\end{thebibliography}



\end{document}